\documentclass{article}

\PassOptionsToPackage{numbers,sort&compress}{natbib}

\usepackage[dblblindworkshop, final]{neurips_2026}

\usepackage[utf8]{inputenc} 
\usepackage[T1]{fontenc}    
\usepackage{hyperref}       
\usepackage{url}            
\usepackage{booktabs}       
\usepackage{amsfonts}       
\usepackage{nicefrac}       
\usepackage{microtype}      
\usepackage{xcolor}         
\usepackage{amsmath}
\usepackage{bm}
\usepackage{amssymb}
\usepackage{graphicx}
\usepackage{soul}
\usepackage{threeparttable}
\usepackage{multirow}
\usepackage{wrapfig}
\usepackage{hyphenat}
\usepackage[table]{xcolor}

\newcommand{\best}[1]{\textcolor{red}{\textbf{#1}}}
\newcommand{\snd}[1]{\textcolor{green!60!black}{\textit{#1}}}
\title{Not All Retained Clients Are Equal: Client Selection for Federated Unlearning}
\workshoptitle{NewInML}
\makeatletter
\renewcommand{\@noticestring}{%
NewInML Workshop at the 40th Conference on Neural
Information Processing Systems (NeurIPS 2026),
Paris, France, 2026.%
}
\makeatother

\author{%
  Thanh Linh Nguyen, Quoc-Viet Pham \\
  School of Computer Science and Statistics,
  Trinity College Dublin, Ireland \\
  \texttt{\{tnguyen3,viet.pham\}@tcd.ie}
}

\newcommand{\Iset}{\mathcal{I}}                 
\newcommand{\Itar}{\mathcal{I}_{\mathrm{tar}}}  
\newcommand{\Iret}{\mathcal{I}_{\mathrm{ret}}}  
\newcommand{\Kset}{\mathcal{K}}                 
\newcommand{\Sset}[1]{\mathcal{S}^{(#1)}}       
\newcommand{\Nc}{N}                             
\newcommand{\kc}{k}                             
\newcommand{\Dloc}[1]{\mathcal{D}_{#1}}         
\newcommand{\nloc}[1]{n_{#1}}                   
 \newcommand{\wun}{\w_{\mathrm{unlearn}}}

\newcommand{\lossw}{\ell}                       
\newcommand{\Floc}[1]{F_{#1}}                   
\newcommand{\Fglob}{F}                          
\newcommand{\Fret}{F_{\mathrm{ret}}}            
\newcommand{\Ftar}{F_{\mathrm{tar}}}            

\newcommand{\wretrain}{\w^{\star}_{\mathrm{retrain}}}
\newcommand{\w}{\bm{w}}

\newcommand{\dimd}{d}                           
\newcommand{\dimL}{d_{\mathrm{cls}}}            
 
\newcommand{\wt}[1]{\w^{(#1)}}              
\newcommand{\Tlearn}{T_{\mathrm{learn}}}        
\newcommand{\Tunlearn}{T_{\mathrm{unlearn}}}    
 
\newcommand{\gfull}[1]{\boldsymbol{g}_{#1}}             
\newcommand{\clw}[2]{\boldsymbol{\theta}^{\mathrm{cls},(#2)}_{#1}}  
\newcommand{\clwg}[1]{\boldsymbol{\theta}^{\mathrm{cls},(#1)}}      
 
\newcommand{\salign}[1]{s^{(#1)}}               
\newcommand{\pdi}{\mathit{pd}}                  
\newcommand{\aind}[1]{a^{(#1)}}                 
\newcommand{\phiw}[1]{\phi^{(#1)}}              
\newcommand{\ncos}{\widehat{\cos}}              

\newcommand{\Tclp}{\mathcal{T}_{\mathrm{CLP}}}  
\newcommand{\FGN}[1]{\mathrm{FGN}^{(#1)}}
\newcommand{\aggw}[1]{p^{(#1)}}                 
\newcommand{\tauclp}{\tau_{\mathrm{CLP}}}
 
\newcommand{\toler}{\delta}                     

\definecolor{MyDarkGray}{cmyk}{0,0,0,0.80}
\newtheorem{definition}{Definition}

\newcommand{\itar}{\mathcal{I}_{\mathrm{tar}}}
\newcommand{\iret}{\mathcal{I}_{\mathrm{ret}}}
\newcommand{\proposed}{{\text{\fontfamily{qcr}\selectfont PDF}}}

\newcommand{\reverse}{{\text{\fontfamily{qcr}\selectfont Rev}}}

\newcommand{\cucoh}
{{\text{\fontfamily{qcr}\selectfont high-Coh}}}
\newcommand{\clps} {{\text{\fontfamily{qcr}\selectfont CLPs}}}
\newcommand{\clp}{{\text{\fontfamily{qcr}\selectfont CLP}}}
\newcommand{\nonclp}{{\text{\fontfamily{qcr}\selectfont non-CLP}}}

\begin{document}

\maketitle

\begin{abstract}
    Federated unlearning (FU) aims to remove a designated target client's influence from a trained model without the prohibitive cost of retraining from scratch. Existing methods largely treat unlearning as a post-hoc, stateless operation, overlooking that client influence is path-dependent: contributions made during training, particularly during critical learning periods (\clps), can have disproportionate and persistent effects. Moreover, knowledge overlap and intermittent client availability make relying on all retained clients inefficient and impractical. We introduce {\proposed}, a path-dependent client-selection framework based on the insight that retained clients are not equally valuable for unlearning. During training, {\proposed} maintains lightweight, target-agnostic, \clp-weighted histories summarizing each client's accumulated update direction and global-model alignment. Upon an unlearning request, these histories identify and rank retained clients that are globally informative yet weakly coupled to the target, enabling retraining-based unlearning with only the top-$k$ clients over a small number of rounds. Experiments on FMNIST, CIFAR-10, and CIFAR-100 under IID and non-IID settings show that {\proposed} achieves effective forgetting and strong retained utility with substantially fewer retained clients. In particular, {\proposed} achieves $0\%$ forgetting accuracy with retraining-like MIA AUC on non-IID CIFAR-10, improves retained accuracy by up to $6.91$ percentage points over the strongest competing FU baseline on CIFAR-100, and reduces communication by approximately $69\%$ and computation by $57$--$61\%$ relative to Retrain.
\end{abstract}

\section{Introduction}
As federated learning (FL) gains traction in privacy-sensitive and highly regulated domains (e.g., healthcare~\cite{theophanous2026international, hanser2025data}), deployed systems must accommodate not only privacy-preserving model training but also the post-training removal of client data influence--whether to comply with legal obligations, such as the right to be forgotten under the GDPR~\citep{gdpr}, or to mitigate impacts of contributions from corrupted, low-quality, or malicious clients. These operational and regulatory demands have given rise to \emph{federated unlearning} (FU), which aims to eliminate the influence of designated target clients from a trained global model while preserving the knowledge contributed by the remaining clients. 

Existing FU methods, including retraining-based approaches~\citep{liu2021federaser, zhang2023fedrecovery, xiong2023exact} and model-manipulation approaches~\citep{10810730, khalil2025not, zhong2025unlearning, gong2026fedortho, nguyen2026computation, weng2026gdfa}, face three major limitations in addressing client-level unlearning.
%
First, they overlook how client influence develops throughout the training stage. In FL, client participation and update directions vary across rounds; therefore, a client's influence on the final model depends not only on \textit{whether} it participates, but also on \textit{when} it participates. This temporal structure is particularly important during critical learning periods ({\clps}), in which client updates can have disproportionately persistent effects on the learned representations, optimization trajectory, and final model performance~\citep{achille2018critical, kleinman2024critical, yan2023criticalfl, yan2022seizing}. Moreover, under heterogeneous data, changing participation, or adaptive optimization, critical phases may emerge beyond a fixed early-training window~\citep{huang2025pa3fed, wang2026gr}. Second, retraining-based methods incur substantial computation, communication, and storage overhead because they reconstruct or recalibrate the model using all retained clients. Their effectiveness therefore depends on continuous availability of clients after training, an assumption that may not hold in practical FL systems with intermittent or voluntary participation. Third, efficient model-manipulation methods often face a pronounced trade-off between target forgetting efficacy and retained-model utility~\citep{gong2026fedortho, chundawat2024conda, weng2026gdfa}. Achieving stronger forgetting may excessively perturb knowledge shared with retained clients, whereas preserving utility may leave residual target influence. They may also rely on expensive server-side operations or require future unlearning targets to be anticipated during the original training stage~\citep{chundawat2024conda, gong2026fedortho, zhong2025unlearning, 10810730, nguyen2026computation}.

We argue that these limitations share a common cause: existing FU methods are largely model layer-, parameter-, or protocol-centric and treat unlearning as a stateless operation, overlooking the impacts of heterogeneous influence that clients accumulate over training. We instead formulate client-level unlearning as an inherently \emph{path-dependent} client-selection problem. Inspired by path dependence in economics~\citep{antonelli1997economics}, we posit that a client's influence is shaped not only by its final contribution, but also by when that influence emerged, particularly during {\clps}, and by how strongly it became coupled with other clients. Thus, effective unlearning should select retained clients according to their historical alignment with the global model and their dependence on the target clients, rather than uniformly or solely by availability. Building on this insight, we ask the following question: \textbf{\textit{How can the path-dependent influence during the learning stage be exploited to improve the forgetting efficacy, retained utility, and efficiency of client-level FU?}}

Therefore, we propose a \underline{\textbf{P}}ath-\underline{\textbf{D}}ependent client selection framework for \underline{\textbf{F}}U, named {\proposed}. During training, {\proposed} maintains lightweight, target-agnostic histories that capture each client’s alignment with the global optimization trajectory, its historical update direction with other clients, and its participation during {\clps}. Upon an unlearning request, these histories are used to estimate each retained client’s coherence with the target set, and the resulting signals are combined into a unified ranking. The ranking reflects our central hypothesis: \emph{retained clients are not equally useful for unlearning}. \textbf{Clients whose updates consistently supported the global model, particularly during critical rounds, are valuable for preserving retained knowledge, whereas clients strongly coupled with the targets may carry overlapping information that hinders selective forgetting.} {\proposed} therefore performs retraining-based unlearning using only the top-$k$ clients that are globally informative yet weakly coupled with the target set, reducing participation, communication, and computation while balancing target forgetting and retained-data generalization. Our contributions are summarized as follows:
\begin{itemize}
\item We introduce a \emph{path-dependent perspective} on client-level FU, showing that retained clients can differ in their value for unlearning because their influence depends on when and how they contributed during training and on their historical dependence on the target clients.

\item  We propose {\proposed}, a {\clp}-aware client-selection framework that maintains lightweight, target-agnostic training histories and request-conditioned target coherence to rank retained clients, enabling retraining-based unlearning with only the top-$k$ clients.

\item Experiments on FMNIST, CIFAR-10, and CIFAR-100 across IID and non-IID settings show that {\proposed} achieves effective forgetting and strong retained utility while substantially reducing client participation, communication, and computation. Ablations further validate the proposed ranking and {\clp}-aware design.
\end{itemize}

\section{Related work}
\textbf{Machine and Federated Unlearning.} Machine unlearning aims to remove the influence of designated training data without full retraining~\cite{cao2015towards}. Existing methods modify either the training process or the learned model through data partitioning, parameter transformation, pruning, or cached training states~\cite{golatkar2020eternal,jia2023model,brophy2021machine}. Such methods generally assume access to raw data, intermediate training states, or sufficient retraining budget---assumptions that fail in FL, where the server cannot access local datasets and clients may be intermittently available~\cite{mcmahan2017communication}. FU extends this objective to client-, sample-, or class-level removal~\cite{romandini2024federated,11415662}. Client-level FU removes the entire contribution of designated clients, encompassing potentially many samples and classes whose influence has accumulated across multiple training rounds. Its methods either approximate retraining from historical updates cached during training~\cite{liu2021federaser,zhang2023fedrecovery, xiong2023exact}, or suppress target influence through parameter, layer, gradient, or representation manipulation~\cite{chundawat2024conda,khalil2025not,gong2026fedortho,nguyen2026computation,pan2025federated, weng2026gdfa}.

\textbf{Critical Learning Periods.} {\clps} are training intervals during which updates exert disproportionately persistent effects on learned representations and final performance~\cite{achille2018critical,kleinman2024critical}. In FL, analogous critical rounds emerge, reflecting the temporally non-uniform nature of client influence~\cite{yan2022seizing,yan2023criticalfl}. Existing {\clp}-aware methods leverage these dynamics to enhance client participation, improve aggregation strategies, or boost training efficiency~\cite{yan2023criticalfl,huang2025pa3fed,wang2026gr}.

\textbf{Discussion.} Existing FU methods utilize training artifacts for post-hoc correction, yet they treat all training rounds and retained clients as equally relevant, overlooking the temporally non-uniform influence. Concurrently, {\clp}-aware FL methods explicitly model these temporal dynamics, but their focus remains strictly on improving model training, leaving their implications for subsequent influence removal unaddressed. This raises an underexplored question: can the temporal patterns captured in the training stage serve as a useful signal for identifying which retained clients are most valuable for unlearning? {\proposed} investigates this question by incorporating {\clp}-weighted signals into client-level FU, exploring whether such temporal histories can effectively guide client selection.

\section{Methodology}

\subsection{System Model}
\label{sec:system-model}

\paragraph{Federated setting.}
We consider an FL system comprising a central server and $\Nc$ clients indexed by $\Iset=\{1,\ldots,\Nc\}$. Each $i\in\Iset$ holds a private dataset $\Dloc{i}=\{(x_j^{(i)},y_j^{(i)})\}_{j=1}^{\nloc{i}}$ of $\nloc{i}=|\Dloc{i}|$ samples, where $(x_j^{(i)},y_j^{(i)})\in\mathcal{X}\times\mathcal{Y}$. Client datasets are record-disjoint and contain $n=\sum_{i\in\Iset}\nloc{i}$ samples in total. All clients share a common model architecture parameterized by $\w\in\mathbb{R}^{\dimd}$, where $\dimd$ is the model dimension. Let $\lossw(\w;x,y)\in\mathbb{R}_{\geq 0}$ be the sample-wise loss. The local empirical objective of client $i$ is
\begin{equation}
\Floc{i}(\w)=\frac{1}{\nloc{i}}\sum_{j=1}^{\nloc{i}}
\lossw\!\left(\w;x_j^{(i)},y_j^{(i)}\right).
\label{eq:local-objective}
\end{equation}
 With $\pi_i=\nloc{i}/n, \sum_{i\in\Iset}\pi_i=1$, the global empirical objective is $
\Fglob(\w)=\sum_{i\in\Iset}\pi_i\Floc{i}(\w)$.
\paragraph{Federated training.}
Training proceeds for $\Tlearn$ communication rounds using FedAvg~\cite{mcmahan2017communication}. At round $t$, the server samples a participating subset $\Sset{t}\subseteq\Iset$ with
$|\Sset{t}|=M\leq\Nc$ uniformly at random without replacement, and broadcasts $\wt{t-1}$ to $\Sset{t}$. Participation is recorded by $\aind{t}_i=\mathbf{1}\{i\in\Sset{t}\}\in\{0,1\}$. Each $i \in \Sset{t}$ performs $E$ local gradient steps with local learning rate $\eta_{\ell,t}>0$:
\begin{align}
\w_i^{(t,0)}&=\wt{t-1}, &
\w_i^{(t,e)}&=\w_i^{(t,e-1)}
-\eta_{\ell,t}\nabla\Floc{i}\!\left(\w_i^{(t,e-1)}\right),
\quad e=1,\ldots,E.
\label{eq:local-update}
\end{align}

We denote the resulting local-update map by $h_i^{(t)}(\wt{t-1})=\w_i^{(t,E)}$. With data-proportional weight $\aggw{t}_i=\nloc{i}/\sum_{m\in\Sset{t}}\nloc{m}$, satisfying $\sum_{i\in\Sset{t}}\aggw{t}_i=1$, the server updates
\begin{equation}
\wt{t}=(1-\gamma_t)\wt{t-1}
+\gamma_t\sum_{i\in\Sset{t}}\aggw{t}_i
h_i^{(t)}(\wt{t-1}),
\label{eq:server-update}
\end{equation}
where $\gamma_t\in(0,1]$ is the server mixing rate and standard FedAvg uses $\gamma_t=1$. We write $\gfull{i}^{(t)}=\nabla\Floc{i}(\wt{t-1})$ for client $i$'s gradient. After $\Tlearn$ rounds the server obtains the trained global model $\wt{\Tlearn}$, together with the training history which is recorded during training and is target-agnostic: it depends on neither $\Itar$ nor any subsequent unlearning request (see details in Sec.~\ref{sec:selection}).

\subsection{Problem Definition}
\label{sec:problem}

We study client-level FU as a two-stage process: a learning stage that produces a trained global model, followed by an unlearning stage that removes the influence of a designated target-client set.

\paragraph{Unlearning request.}
After training, a nonempty target-client set $\Itar\subsetneq\Iset$ is designated for removal, and the retained set is $\Iret=\Iset\setminus\Itar$, with $|\Iret|+|\Itar|=\Nc$. Let $n_{\mathrm{ret}}=\sum_{i\in\Iret}\nloc{i}$ and $n_{\mathrm{tar}}=\sum_{o\in\Itar}\nloc{o}$.The retained-client and
target-client objectives are as follows:
\begin{align}
\Fret(\w)
&=\sum_{i\in\Iret}\frac{\nloc{i}}{n_{\mathrm{ret}}}\Floc{i}(\w),
&
\Ftar(\w)
&=\sum_{o\in\Itar}\frac{\nloc{o}}{n_{\mathrm{tar}}}\Floc{o}(\w).
\label{eq:ret-target-objectives}
\end{align}

\paragraph{Retraining reference and unlearning objective.}
Let $\mathcal{A}$ denote the original FL training algorithm, including its initialization, client sampling, and optimization randomness $\xi$. The retraining reference is $\wretrain=\mathcal{A}(\{\Dloc{i}\}_{i\in\Iret};\xi)$, obtained by rerunning the training procedure from scratch using only retained clients. Given $\wt{\Tlearn}$, $\Itar$, training history, an unlearning algorithm returns $\wun$. We seek retraining-like retained and target behavior:
\begin{equation}
\Fret(\wun)-\Fret(\wretrain)\leq\toler,
\qquad
\left|\Ftar(\wun)-\Ftar(\wretrain)\right|\leq\varepsilon_f,
\label{eq:unlearning-criteria}
\end{equation}
where $\toler,\varepsilon_f\geq0$ denote the retained-utility and
forgetting tolerances, respectively.

\begin{figure*}[t!]
	\centering
	\includegraphics[width=\linewidth]{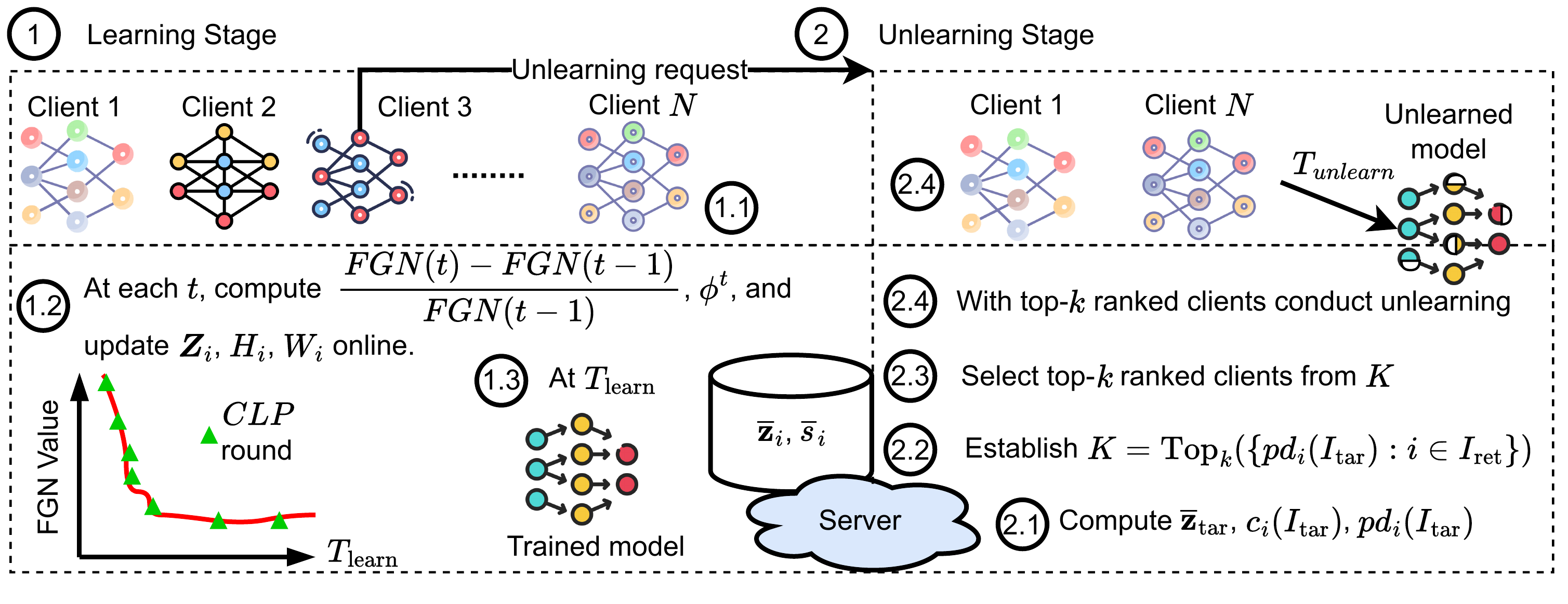}
	\caption{{\proposed} framework with workflow.}
    \label{fig:framework}
\end{figure*}

\section{Path-Dependent Client Selection for Unlearning}
\label{sec:selection}

{\proposed} operates in two stages as shown in Fig.~\ref{fig:framework}. During federated training, the server detects {\clps} and maintains target-agnostic, {\clp}-weighted histories of each client's update direction and global alignment. Once an unlearning request reveals the target set $\Itar$, these histories are used to measure each retained client's coherence with the target clients and rank its suitability for unlearning. This design neither requires the target set to be known during training nor stores per-round model snapshots. To reduce storage and computation, all statistics are computed on the classification layer, which is sensitive to client-specific and non-IID variation~\cite{wang2025fedcm,li2023no,oh2022fedbabu,yi2024fedssa,luo2021no}.

\subsection{Learning stage}
\label{subsec:clp_dectection}

\begin{wrapfigure}{r}{0.47\textwidth}
\vspace{-\intextsep}
\centering
\includegraphics[width=0.47\textwidth]{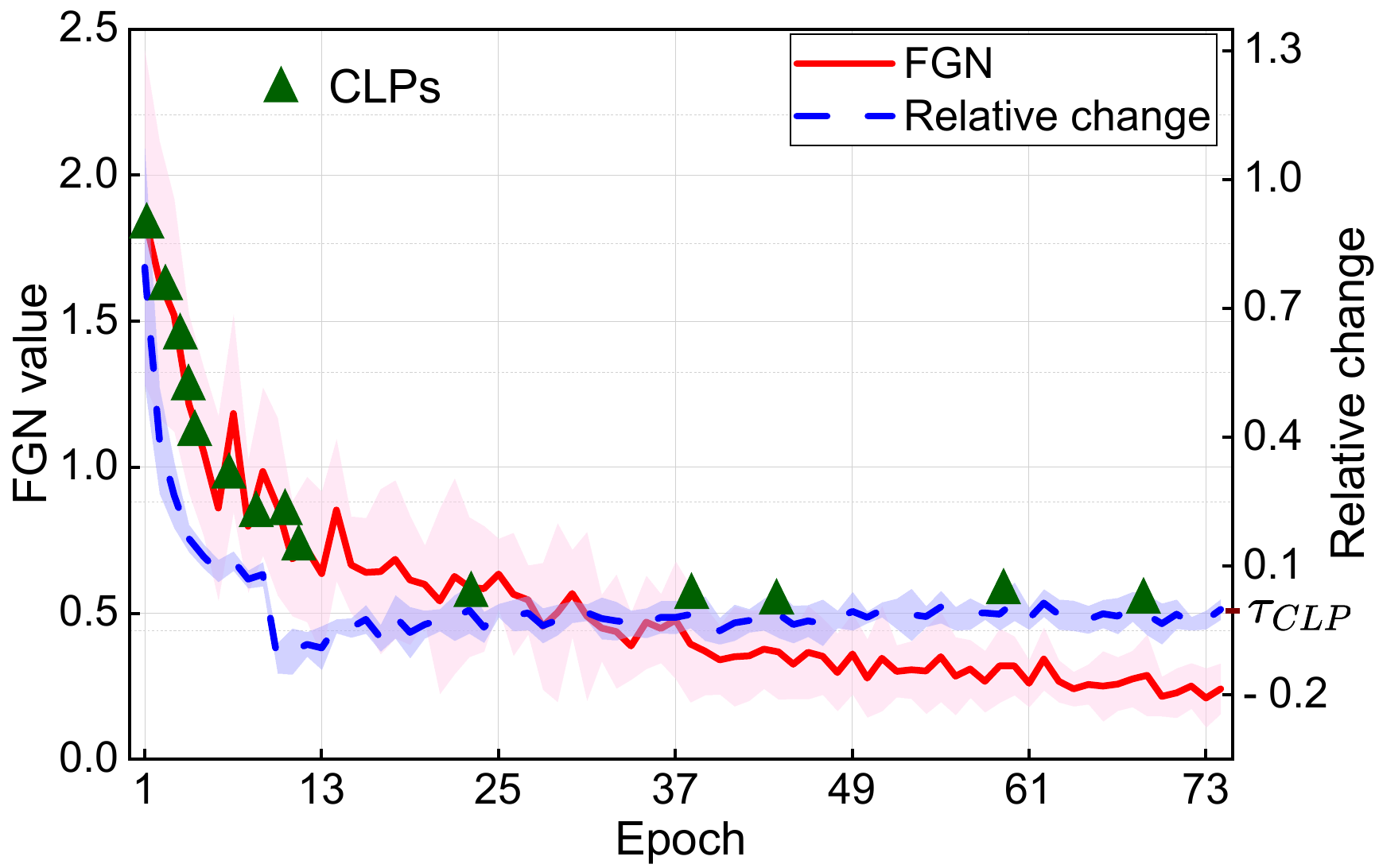}
\caption{Illustration of {\clp} detection using the FGN proxy on FMNIST~\cite{xiao2017fashion}, where the green double-headed arrows and triangles indicate the detected {\clps}.}
\label{fig:clp_probing}
\vspace{-2\intextsep}
\end{wrapfigure}

\paragraph{Critical learning period detection.} The onset and duration of {\clps} vary across training trajectories and are not confined to a fixed early interval (see Fig.~\ref{fig:clp_probing})~\cite{huang2025pa3fed}. Let $\Tclp=\bigcup\nolimits_{q=1}^{Q}\Tclp^{(q)}\subseteq\{1,\dots,\Tlearn\}$ denote the set of detected critical rounds, where $Q$ is the number of identified critical periods.  To detect {\clps} online, the server employs the federated gradient norm (FGN) proxy~\cite{yan2023criticalfl}. Given the round-$t$ local gradient $\gfull{i}^{(t)}$, one local training step produces the first-order loss change: $\Delta\Floc{i}^{(t)}
   =\Floc{i}\!\big(\wt{t-1}-\eta_{\ell,t}\,\gfull{i}^{(t)}\big)
    -\Floc{i}\!\big(\wt{t-1}\big) 
   \approx -\,\eta_{\ell,t}\big\|\gfull{i}^{(t)}\big\|_2^{2}
$. Accordingly, the round-level loss-decrease proxy is defined as $\FGN{t}\triangleq\,\eta_{\ell,t}\!\!\sum_{i\in\Sset{t}}\!\aggw{t}_i\big\|\gfull{i}^{(t)}\big\|_2^{2}$. A round $t$ is flagged as critical (i.e.\ $t\in\Tclp$), when $\frac{\FGN{t}-\FGN{t-1}}{\FGN{t-1}}\; \ge \;\tauclp,$ where $\tauclp > 0$ is an empirically selected sensitivity threshold \citep{yan2023criticalfl,huang2025pa3fed}. Evaluating this criterion throughout training enables the detection of both early and late critical periods (see Fig.~\ref{fig:clp_probing}).

Throughout, $\clw{i}{t}\in\mathbb{R}^{\dimL}$ denotes the vectorized classification-layer parameters of client $i$ after local training at round $t$, and $\clwg{t}\in\mathbb{R}^{\dimL}$ denotes the corresponding aggregated global parameters, where $\dimL$ is the classification-layer dimension.
\begin{definition}[Server Alignment]
\label{def:align}
For each $i\in\Sset{t}$, we define client and global classification-layer updates as: $
\boldsymbol{\delta}^{\mathrm{cls},(t)}_i
=
\clw{i}{t}-\clwg{t-1}, \quad
\boldsymbol{\delta}^{\mathrm{cls},(t)}_g
=
\clwg{t}-\clwg{t-1}
$, respectively. Therefore, the round-level server alignment of client $i$ is
\begin{equation}
\salign{t}_i
=
\ncos\!\left(
\boldsymbol{\delta}^{\mathrm{cls},(t)}_i,
\boldsymbol{\delta}^{\mathrm{cls},(t)}_g
\right),
\label{eq:align}
\end{equation}
where $
\ncos(\bm{a},\bm{b})
=
\frac{1}{2}\!\left(
1+
\frac{\langle\bm{a},\bm{b}\rangle}
{(\|\bm{a}\|_2+\varepsilon_0)(\|\bm{b}\|_2+\varepsilon_0)}
\right)$ 
is the normalized cosine similarity and $\varepsilon_0>0$ ensures numerical stability. A larger $\salign{t}_i$ indicates stronger consistency between client $i$'s local update and the realized global update at round $t$.
\end{definition}

\begin{definition}[\clp-Aware Historical Accumulation]
\label{def:history}
To emphasize contributions made during critical rounds, we assign round $t$ the temporal weight:
$
\phiw{t}
=
\begin{cases}
1+\lambda, & t\in\Tclp,\\[2pt]
1, & \text{otherwise,}
\end{cases}
$
where $\lambda \ge 0$ controls the additional weight assigned to {\clp} contributions. For each participating client, let
\begin{equation}
\widehat{\boldsymbol{\delta}}^{\mathrm{cls},(t)}_i
=
\frac{\boldsymbol{\delta}^{\mathrm{cls},(t)}_i}
{\|\boldsymbol{\delta}^{\mathrm{cls},(t)}_i\|_2+\varepsilon_0}
\label{eq:normalized-client-update}
\end{equation}
denote its normalized update direction. The server initializes
$\boldsymbol{Z}^{(0)}_i=\boldsymbol{0}$, $H^{(0)}_i=0$, and $W^{(0)}_i=0$, and recursively updates
\begin{align}
\boldsymbol{Z}^{(t)}_i
=
\boldsymbol{Z}^{(t-1)}_i
+
\phiw{t}\aind{t}_i
\widehat{\boldsymbol{\delta}}^{\mathrm{cls},(t)}_i, \;
H^{(t)}_i
=
H^{(t-1)}_i
+
\phiw{t}\aind{t}_i\salign{t}_i, \;
W^{(t)}_i
=
W^{(t-1)}_i
+
\phiw{t}\aind{t}_i.
\end{align}
At $\Tlearn$, the accumulated update direction and historical server alignment of client $i$ are as follows:
\begin{align}
\boldsymbol{z}_i
=
\frac{\boldsymbol{Z}^{(\Tlearn)}_i}
{W^{(\Tlearn)}_i+\varepsilon_0}, \quad
\bar{\boldsymbol{z}}_i
=
\frac{\boldsymbol{z}_i}
{\|\boldsymbol{z}_i\|_2+\varepsilon_0}, \quad
\bar{s}_i
=
\frac{H^{(\Tlearn)}_i}
{W^{(\Tlearn)}_i+\varepsilon_0}.
\end{align}
Here, $\bar{\boldsymbol{z}}_i$ summarizes the dominant direction of client $i$'s historical updates, emphasizing contributions made during {\clps}, while $\bar{s}_i$ measures its average historical alignment with the global trajectory. These target-agnostic statistics require one $\dimL$-dimensional vector and two scalars per client.
\end{definition}

\subsection{Unlearning stage} 
Once an unlearning request reveals the target-client set $\Itar$, the server uses the cached learning-stage histories to quantify each retained client's dependence on the target set.

\begin{definition}[Target Coherence]
\label{def:coh}
The aggregate historical direction of the target clients is
$
\bar{\boldsymbol{z}}_{\mathrm{tar}}
=
\frac{\sum_{o\in\Itar}\bar{\boldsymbol{z}}_o}
{\left\|\sum_{o\in\Itar}\bar{\boldsymbol{z}}_o\right\|_2+\varepsilon_0}.
$
For each retained client $i\in\Iret$, its target coherence is defined by:
\begin{equation}
c_i(\Itar)
=
\ncos\!\left(
\bar{\boldsymbol{z}}_i,
\bar{\boldsymbol{z}}_{\mathrm{tar}}
\right).
\label{eq:coh}
\end{equation}
A larger $c_i(\Itar)$ indicates stronger alignment between the historical update directions of client $i$ and the target set, whereas $1-c_i(\Itar)$ measures historical target decoupling. \textbf{Selecting clients with larger decoupling reduces the tendency of retained-client updates to restore target-related knowledge during the subsequent unlearning process} (see Tab.~\ref{tab:ablation_proposed}).
\end{definition}

\begin{definition}[Path-Dependence Unlearning Score]
\label{def:pd}
For each retained client $i\in\Iret$, we define the path-dependence score as
\begin{equation}
\pdi_i(\Itar)
=
\beta\bar{s}_i
+
(1-\beta)\bigl[1-c_i(\Itar)\bigr].
\label{eq:pd}
\end{equation}
The parameter $\beta\in[0,1]$ balances historical global alignment and target decoupling: larger values emphasize clients whose updates consistently followed the global learning trajectory, thereby favoring retained-model utility preservation, whereas smaller values emphasize clients whose historical directions are weakly coupled with those of the target clients, thereby favoring target forgetting.
\end{definition}
A larger $\pdi_i(\Itar)$ therefore favors retained clients that are both globally aligned and historically decoupled from the target set, with contributions during detected {\clps} emphasized through $\phiw{t}$. The server selects the $\kc$ highest-scoring retained clients:
\begin{align}
\Kset
=
\mathrm{Top}_{\kc}
\left(
\left\{
\pdi_i(\Itar):i\in\Iret
\right\}
\right), \text{ with }
\kc\leq|\Iret|.
\label{eq:topk}
\end{align}
Only clients in $\Kset$ participate in the subsequent $\Tunlearn$ unlearning rounds. The persistent learning-stage cache therefore requires $\mathcal{O}(\Nc\dimL)$ space, while unlearning-stage communication and computation scale with $\kc$ selected clients rather than all $|\Iret|$ retained clients.

\section{Experimental evaluation}

\subsection{Experiment setup}
\textbf{Unlearning scenario.} We simulate Byzantine attacks via label flipping, where a single client's training labels are maliciously manipulated, triggering an unlearning request for that client~\cite{zhong2025unlearning}.

\textbf{Hyperparameter, dataset, and model.} We simulate an FL system using FedAvg~\cite{mcmahan2017communication} for global model training with $N=64$ clients. Unless otherwise stated, $M=16$ clients are sampled per training round, with $\eta_{l,t}=0.01$, batch size $32$, {\clp} detection threshold $\tau_{\clp}=0.01$. We consider both IID and Dirichlet-based non-IID data partitions with $\alpha \in \{1, 0.1\}$,  where a smaller $\alpha$ indicates stronger data heterogeneity. For FU, we set $\beta \in \{0.1, 0.5, 0.9\}$ to control the trade-off between retained knowledge preservation and the removal of target-client influence,  with smaller $\beta$ yielding stronger forgetting; $\lambda=0.7$, which adds weights to participation during {\clps}; and $T_{\mathrm{unlearn}} = 20$. We evaluate on FMNIST~\cite{xiao2017fashion}, CIFAR-10, and CIFAR-100~\cite{cifar}. A CNN architecture is used for FMNIST (see Appendix~\ref{subsec:cnn}), while ResNet-18~\cite{he2016deep} is adopted for CIFAR-10 and CIFAR-100.

\textbf{Baselines.} We compare {\proposed} with Retrain, CONDA~\cite{chundawat2024conda, foster2024fast}, FedRecovery~\cite{zhang2023fedrecovery}, FOUL~\cite{nguyen2026computation}, FedEraser~\cite{liu2021federaser}, Exact-Fun (E-F)~\cite{xiong2023exact}, FUSED~\cite{zhong2025unlearning}, and GDFA~\cite{weng2026gdfa} (see details in Appendix~\ref{subsec:baseline}).

\textbf{Evaluation metric.} We evaluate unlearning efficacy, efficiency, and fidelity, using the following active testing  metrics~\cite{11359714, nguyen2026computation,zhong2025unlearning, gu2024ferrari, chundawat2024conda, weng2026gdfa}:
\begin{itemize}
    \item \textbf{FA \& RA} (\%): Forgetting accuracy (FA) measures predictive accuracy on the target-client data, where lower FA indicates stronger removal of target-client influence. Retained accuracy (RA) measures accuracy on the retained data, where higher RA indicates better preservation of retained knowledge.
        
    \item \textbf{MIA AUC}  (\%): We assess residual privacy leakage using membership inference attacks (MIA)~\cite{shokri2017membership}. AUC measures member--non-member distinguishability across decision thresholds, with $50\%$ corresponding to random discrimination. We use Retrain as the reference and report the MIA-AUC gap to Retrain, where a smaller gap indicates more retraining-like privacy behavior.

    \item \textbf{Comm} (GB) \& \textbf{Comp} (PFLOPs): Total data transmitted and total floating-point operations (FLOPs), respectively, each aggregated over training and unlearning. We include training because several learning-to-unlearn baselines modify the training protocol, so measuring the unlearning stage alone would understate their cost.


    \item \textbf{Space Complexity}: We measure the system-wide memory required to store model checkpoints, historical statistics, and intermediate model updates during training and unlearning.
\end{itemize}

\begin{table*}[t]
\centering
\caption{Performance comparison between {\proposed} and other FU methods on three datasets under $\alpha \in \{0.1, 1.0\}$. Subscripts on {\proposed} denote $\beta \in \{0.1, 0.5, 0.9\}$, which controls the forgetting--utility trade-off: \textbf{$0.1$ is forgetting-oriented, $0.5$ balances forgetting and utility preservation, and $0.9$ is utility-preservation-oriented}. All variants use $k=25$ selected unlearning clients. Values in parentheses are deltas relative to Retrain. For FA, RA, and MIA AUC, \best{red} marks the best and \snd{green} the second-best method across all unlearning methods (Retrain is the reference and is excluded), ranked by FA ($\downarrow$), RA ($\uparrow$), and $|\Delta\mathrm{MIA}|$ ($\downarrow$).}
\label{tab:client_unlearning_results}
\renewcommand{\arraystretch}{1.05}
\setlength{\tabcolsep}{5pt}
\resizebox{\textwidth}{!}{%
\begin{tabular}{
l|l|c|
ccccccc|
>{\columncolor{gray!10}}c
>{\columncolor{gray!10}}c
>{\columncolor{gray!10}}c
}
\toprule
Metric & $\alpha$ & Retrain & FUSED & CONDA & E-F & FedEraser & FOUL & FedRecovery & GDFA
& {\proposed}$_{0.1}$ & {\proposed}$_{0.5}$ & {\proposed}$_{0.9}$ \\
\midrule

\multicolumn{13}{c}{\textit{FMNIST-CNN}} \\
\midrule
\multirow{2}{*}{FA}
& 0.1 & 0.44 & 0.44 ($+$0.00) & 0.70 ($+$0.26) & 0.56 ($+$0.12) & 1.11 ($+$0.67) & 0.75 ($+$0.31) & 0.60 ($+$0.16) & \snd{0.28 ($-$0.16)} & \best{0.00 ($-$0.44)} & 0.72 ($+$0.28) & 0.44 ($+$0.00) \\
& 1.0 & 0.92 & 0.82 ($-$0.10) & 0.94 ($+$0.02) & 0.45 ($-$0.47) & 0.53 ($-$0.39) & 0.77 ($-$0.15) & 0.87 ($-$0.06) & \best{0.14 ($-$0.78)} & \snd{0.28 ($-$0.64)} & \snd{0.28 ($-$0.64)} & 0.92 ($+$0.00) \\
\addlinespace[1.5pt]
\multirow{2}{*}{RA}
& 0.1 & 88.20 & \best{87.14 ($-$1.06)} & 85.17 ($-$3.02) & 81.89 ($-$6.31) & 61.04 ($-$27.16) & 85.60 ($-$2.60) & 85.22 ($-$2.98) & 81.83 ($-$6.37) & 82.60 ($-$5.60) & 85.97 ($-$2.23) & \snd{86.89 ($-$1.31)} \\
& 1.0 & 90.45 & \snd{89.60 ($-$0.85)} & 89.40 ($-$1.05) & 85.94 ($-$4.50) & 76.42 ($-$14.03) & \best{89.64 ($-$0.81)} & \best{89.64 ($-$0.81)} & 88.25 ($-$2.20) & 89.34 ($-$1.11) & 89.35 ($-$1.10) & 89.55 ($-$0.90) \\
\addlinespace[1.5pt]
\multirow{2}{*}{\shortstack[l]{MIA\\AUC}}
& 0.1 & 50.32 & \snd{50.05 ($-$0.27)} & \best{50.11 ($-$0.20)} & 49.84 ($-$0.48) & 49.92 ($-$0.40) & 49.71 ($-$0.61) & 50.01 ($-$0.31) & 50.00 ($-$0.32) & 50.01 ($-$0.31) & 50.00 ($-$0.32) & 49.91 ($-$0.41) \\
& 1.0 & 50.23 & 49.79 ($-$0.44) & 50.09 ($-$0.15) & \best{50.25 ($+$0.02)} & 50.08 ($-$0.15) & 49.82 ($-$0.41) & 50.16 ($-$0.07) & 50.17 ($-$0.06) & 50.13 ($-$0.10) & \best{50.25 ($+$0.02)} & \snd{50.19 ($-$0.04)} \\
\addlinespace[1.5pt]
\multirow{1}{*}{Comm}
& 0.1/1.0 & 15.62 & 4.11 & 3.97 & 7.82 & 4.36 & 6.08 & 3.97 & 3.98 & 4.79 & 4.79 & 4.79 \\
\addlinespace[1.5pt]
\multirow{2}{*}{Comp}
& 0.1 & 2.09 & 1.01 ($-$1.09) & 0.71 ($-$1.39) & 1.22 ($-$0.87) & 0.72 ($-$1.37) & 1.71 ($-$0.38) & 0.71 ($-$1.39) & 0.71 ($-$1.38) & 0.83 ($-$1.26) & 0.88 ($-$1.22) & 0.89 ($-$1.21) \\
& 1.0 & 2.10 & 1.01 ($-$1.09) & 0.71 ($-$1.39) & 1.23 ($-$0.87) & 0.73 ($-$1.37) & 1.72 ($-$0.38) & 0.71 ($-$1.39) & 0.71 ($-$1.39) & 0.83 ($-$1.27) & 0.83 ($-$1.27) & 0.84 ($-$1.26) \\
\addlinespace[1.5pt]
\midrule

\multicolumn{13}{c}{\textit{CIFAR-10-ResNet}} \\
\midrule
\multirow{2}{*}{FA}
& 0.1 & 6.93 & 6.14 ($-$0.79) & 7.52 ($+$0.59) & 2.21 ($-$4.72) & 4.17 ($-$2.76) & 1.86 ($-$5.07) & 0.77 ($-$6.16) & 4.45 ($-$2.48) & 0.42 ($-$6.51) & \best{0.00 ($-$6.93)} & \snd{0.39 ($-$6.54)} \\
& 1.0 & 0.72 & 1.38 ($+$0.66) & 11.18 ($+$10.46) & 3.93 ($+$3.20) & 7.97 ($+$7.25) & 1.96 ($+$1.24) & 5.87 ($+$5.15) & \best{0.33 ($-$0.39)} & \snd{0.89 ($+$0.17)} & 1.09 ($+$0.37) & 0.94 ($+$0.22) \\
\addlinespace[1.5pt]

\multirow{2}{*}{RA}
& 0.1 & 55.28 & 49.85 ($-$5.43) & 14.68 ($-$40.60) & 23.10 ($-$32.18) & 9.68 ($-$45.60) & 52.10 ($-$3.18) & 37.12 ($-$18.16) & 38.61 ($-$16.67) & 45.22 ($-$10.06) & \snd{54.48 ($-$0.80)} & \best{57.10 ($+$1.82)} \\
& 1.0 & 86.93 & \snd{83.56 ($-$3.38)} & 9.01 ($-$77.93) & 52.40 ($-$34.53) & 24.67 ($-$62.26) & 82.65 ($-$4.28) & 46.83 ($-$40.10) & 69.34 ($-$17.59) & 82.28 ($-$4.65) & 82.84 ($-$4.09) & \best{84.08 ($-$2.85)} \\
\addlinespace[1.5pt]

\multirow{2}{*}{\shortstack[l]{MIA\\AUC}}
& 0.1 & 55.64 & 61.43 ($+$5.79) & 62.98 ($+$7.34) & 50.13 ($-$5.51) & 49.87 ($-$5.78) & 60.73 ($+$5.09) & \best{55.16 ($-$0.49)} & 57.53 ($+$1.89) & 54.62 ($-$1.02) & \snd{54.76 ($-$0.88)} & 56.69 ($+$1.05) \\
& 1.0 & 52.18 & 54.30 ($+$2.12) & 53.38 ($+$1.20) & 49.33 ($-$2.85) & 50.54 ($-$1.65) & 53.11 ($+$0.93) & 51.21 ($-$0.98) & \best{51.82 ($-$0.36)} & 51.74 ($-$0.44) & \snd{51.78 ($-$0.40)} & 51.58 ($-$0.60) \\
\addlinespace[1.5pt]

\multirow{1}{*}{Comm}
& 0.1/1.0 & 845.48 & 217.29 & 214.73 & 423.39 & 235.88 & 329.25 & 214.73 & 215.35 & 259.46 & 259.46 & 259.46 \\
\addlinespace[1.5pt]

\multirow{2}{*}{Comp}
& 0.1 & 47.45 & 22.38 ($-$25.07) & 16.01 ($-$31.45) & 27.69 ($-$19.76) & 16.42 ($-$31.03) & 38.45 ($-$9.00) & 16.01 ($-$31.45) & 16.02 ($-$31.43) & 18.73 ($-$28.72) & 19.30 ($-$28.15) & 20.02 ($-$27.43) \\
& 1.0 & 47.63 & 22.50 ($-$25.14) & 16.10 ($-$31.53) & 27.84 ($-$19.79) & 16.49 ($-$31.14) & 38.66 ($-$8.97) & 16.10 ($-$31.53) & 16.11 ($-$31.52) & 18.70 ($-$28.93) & 18.76 ($-$28.87) & 18.88 ($-$28.75) \\
\addlinespace[1.5pt]
\midrule

\multicolumn{13}{c}{\textit{CIFAR-100-ResNet}} \\
\midrule
\multirow{2}{*}{FA}
& 0.1 & 0.27 & 1.26 ($+$0.99) & 0.86 ($+$0.59) & 1.46 ($+$1.19) & 0.84 ($+$0.56) & 1.50 ($+$1.23) & 0.86 ($+$0.59) & 0.86 ($+$0.59) & \best{0.00 ($-$0.27)} & 0.86 ($+$0.59) & \snd{0.52 ($+$0.25)} \\
& 1.0 & 0.60 & 0.91 ($+$0.30) & 0.54 ($-$0.07) & \snd{0.40 ($-$0.21)} & 1.25 ($+$0.64) & 1.00 ($+$0.40) & 1.44 ($+$0.83) & \best{0.08 ($-$0.52)} & 0.44 ($-$0.16) & 0.91 ($+$0.31) & 0.52 ($-$0.08) \\
\addlinespace[1.5pt]

\multirow{2}{*}{RA}
& 0.1 & 55.17 & 36.34 ($-$18.83) & 4.35 ($-$50.82) & 16.98 ($-$38.20) & 3.98 ($-$51.19) & 42.22 ($-$12.95) & 1.03 ($-$54.14) & 42.15 ($-$13.02) & \snd{48.63 ($-$6.54)} & 48.43 ($-$6.74) & \best{49.13 ($-$6.04)} \\
& 1.0 & 58.92 & 31.83 ($-$27.09) & 2.12 ($-$56.80) & 23.80 ($-$35.12) & 9.10 ($-$49.82) & 39.81 ($-$19.11) & 14.46 ($-$44.46) & 47.50 ($-$11.42) & \best{54.38 ($-$4.54)} & \snd{54.24 ($-$4.68)} & 54.21 ($-$4.71) \\
\addlinespace[1.5pt]

\multirow{2}{*}{\shortstack[l]{MIA\\AUC}}
& 0.1 & 51.44 & 54.02 ($+$2.59) & 55.48 ($+$4.04) & 55.21 ($+$3.77) & \snd{50.71 ($-$0.73)} & 53.09 ($+$1.65) & 49.96 ($-$1.48) & 52.42 ($+$0.98) & \best{52.13 ($+$0.69)} & 52.57 ($+$1.13) & 52.43 ($+$0.99) \\
& 1.0 & 62.15 & \snd{53.91 ($-$8.24)} & 50.20 ($-$11.95) & \best{54.90 ($-$7.25)} & 51.36 ($-$10.79) & 50.49 ($-$11.66) & 50.45 ($-$11.70) & 49.73 ($-$12.42) & 53.03 ($-$9.12) & 52.39 ($-$9.76) & 52.98 ($-$9.17) \\
\addlinespace[1.5pt]

\multirow{1}{*}{Comm}
& 0.1/1.0 & 848.97 & 218.19 & 215.61 & 425.14 & 236.86 & 330.60 & 215.61 & 216.24 & 260.53 & 260.53 & 260.53 \\
\addlinespace[1.5pt]

\multirow{2}{*}{Comp}
& 0.1 & 47.75 & 22.63 ($-$25.11) & 16.22 ($-$31.53) & 28.02 ($-$19.72) & 16.62 ($-$31.13) & 38.93 ($-$8.82) & 16.22 ($-$31.53) & 16.23 ($-$31.52) & 18.83 ($-$28.92) & 18.87 ($-$28.88) & 18.96 ($-$28.78) \\
& 1.0 & 47.77 & 22.59 ($-$25.18) & 16.17 ($-$31.60) & 27.96 ($-$19.81) & 16.57 ($-$31.20) & 38.82 ($-$8.95) & 16.17 ($-$31.60) & 16.18 ($-$31.59) & 18.72 ($-$29.05) & 18.73 ($-$29.04) & 18.75 ($-$29.02) \\
\bottomrule

\end{tabular}}
\end{table*}

\subsection{Main result} As shown in Tab.~\ref{tab:client_unlearning_results}, {\proposed} achieves effective forgetting while preserving retained knowledge using only $25$ clients for unlearning. On FMNIST, {\proposed} remains competitive with the strongest baselines, reaching $86.89\%$ RA under $\alpha=0.1$ while the forgetting-oriented variant achieves $0\%$ FA. Its advantage becomes clearer on the more challenging CIFAR tasks. Under heterogeneous CIFAR-10 (i.e., $\alpha=0.1$), {\proposed}$_{0.9}$ achieves $57.10\%$ RA, exceeding both Retrain ($55.28\%$) and FOUL ($52.10\%$), while {\proposed}$_{0.5}$ combines $54.48\%$ RA with $0\%$ FA. On CIFAR-100, all three variants consistently outperform the strongest competing baseline in RA under both data distributions, reaching up to $49.13\%$ versus $42.22\%$ under $\alpha=0.1$ and $54.38\%$ versus $47.50\%$ under $\alpha=1.0$. \textit{\textbf{These results support our central hypothesis that retained clients differ in their value for unlearning, and that exploiting their path-dependent training histories enables better preservation of retained knowledge}}. The variants further demonstrate the intended role of $\beta$ in controlling the forgetting--utility trade-off. Increasing $\beta$ emphasizes historical global alignment and generally favors utility preservation, whereas smaller $\beta$ emphasizes target decoupling and stronger forgetting. For example, under $\alpha=0.1$, RA increases from $82.60\%$ to $86.89\%$ on FMNIST and from $45.22\%$ to $57.10\%$ on CIFAR-10 as $\beta$ increases from $0.1$ to $0.9$. The weaker variation across $\beta$ on CIFAR-100 suggests that {\proposed} is less sensitive to the trade-off parameter in this setting. Finally, {\proposed} reduces communication by approximately $69\%$ and computation by $57$--$61\%$ relative to Retrain, demonstrating a favorable trade-off among forgetting, retained utility, and unlearning efficiency.

\begin{wraptable}{r}{0.48\textwidth}   
\vspace{-1.5\intextsep}                   
\caption{Space complexity analysis.}
\label{tab:space_complexity}
\begin{threeparttable}
\resizebox{0.3\textwidth}{!}{%
\begin{tabular}{l|c}
\toprule
Method & Space Complexity \\
\midrule
Retrain
& $\mathcal{O}\!\left(P+|\mathcal{I}_{\mathrm{ret}}|\cdot P\right)$ \\
FUSED
& $\mathcal{O}\!\left(P+|\mathcal{I}_{\mathrm{ret}}|\cdot A\right)$ \\
CONDA
& $\mathcal{O}\!\left(N\cdot P+P\right)$ \\
E-F
& $\mathcal{O}\!\left(T_{\mathrm{learn}}\cdot M\cdot P+P\right)$ \\
FedEraser
& $\mathcal{O}\!\left(H\cdot M\cdot P+|\mathcal{I}_{\mathrm{ret}}|\cdot P\right)$ \\
FOUL
& $\mathcal{O}\!\left(P+N\cdot P+N^2\right)$ \\
FedRecovery
& $\mathcal{O}\!\left(T_{\mathrm{learn}}\cdot M\cdot P+P\right)$ \\
GDFA
& $\mathcal{O}\!\left(K_{\text{per}}\cdot P+|\itar|\cdot P\right)$ \\
\midrule
{\proposed}
& $\mathcal{O}\!\left(N\cdot \dimL+\kc\cdot P\right)$ \\
\bottomrule
\end{tabular}%
}
\begin{tablenotes}[flushleft]
\scriptsize
\item[$*$] $P=|\mathbf{w}|$ is the number of full-model parameters.
\item[$\dagger$] $A$ is the number of LoRA unlearning parameters, $A\ll P$.
\item[$\dagger\dagger$]  $\dimL\ll P$.
\item[$\ddagger$] $H$ is the number of FedEraser checkpoints, $H<T_{\mathrm{learn}}$.
\item[$\bullet$] $K_{\text{per}}$ is the number of GDFA perturbed models.
\item[$\diamond$] $\kc$ is the number of retained clients selected by {\proposed}, $\kc<|\iret|$.
\end{tablenotes}
\end{threeparttable}
\vspace{-\intextsep}
\end{wraptable}
As shown in Tab.~\ref{tab:space_complexity}, {\proposed} is space-efficient because its learning-stage cache stores only one accumulated classification-layer update direction and two scalar statistics per client, yielding $\mathcal{O}(N\dimL)$ storage independent of the number of learning rounds, where $\dimL\ll P$. During unlearning, {\proposed} operates on only the selected $\kc$ retained clients, so the model and update tensors instantiated during computation scale with $\mathcal{O}(\kc P)$ rather than all retained clients, where $\kc<|\iret|$. In comparison, Retrain and FUSED involve all retained clients, although FUSED reduces per-client storage and computation through lightweight LoRA parameters. CONDA and FOUL maintain client-wise full-model information, while GDFA constructs and fine-tunes multiple perturbed full-model instances. FedEraser additionally stores checkpointed model histories, whereas E-F and FedRecovery retain full-model information across learning rounds.

\subsection{Ablation study}

As shown in Tab.~\ref{tab:ablation_unlearning_clients_beta}, increasing $\kc$ generally improves retained utility, but the gains diminish as more clients participate. At $\beta{=}0.5$, for example, RA rises from $79.26\%$ at $\kc{=}10$ to $85.97\%$ at $\kc{=}25$, but increases only marginally to $86.40\%$ at $\kc{=}30$. A similar saturation occurs for $\beta{=}0.9$, where increasing $\kc$ from $20$ to $25$ improves RA by $0.9$ percentage points, compared with only $0.1$ percentage points from $25$ to $30$. In contrast, communication and computation increase steadily with $\kc$, revealing a clear utility--efficiency trade-off. MIA AUC remains within $0.41$ percentage points of Retrain throughout the sweep, indicating that this trade-off does not alter retraining-like privacy behavior. The observed variation across $\beta$ is consistent with the forgetting--utility behavior discussed in Tab.~\ref{tab:client_unlearning_results}. We therefore adopt $\kc{=}25$ as the default.
\begin{table*}[h]
\centering
\caption{Ablation study on the number of selected unlearning clients $\kc$ and
knowledge retention--forgetting tradeoff $\beta$ on FMNIST under $\alpha=0.1$
with $\itar={0}$. MIA AUC values in parentheses denote differences from
Retrain ($50.32$). The shaded column denotes the default $\kc=25$ in Tab.~\ref{tab:client_unlearning_results}.}
\label{tab:ablation_unlearning_clients_beta}
\resizebox{0.9\textwidth}{!}{%
\begin{tabular}{c|l|ccc>{\columncolor{gray!12}}ccc}
\toprule
& $\kc$ & $10$ & $15$ & $20$ & \textbf{$25$} & $30$ & $35$ \\
\midrule

\multirow{4}{*}{$\beta=0.1$}
& FA
& 0.00 & 0.00 & 0.00 & 0.00 & 0.28 & 0.28 \\
& RA
& 75.65 & 80.51 & 81.26 & 82.60 & 84.45 & 85.18 \\
& MIA AUC
& 49.96 ($-$0.36)
& 50.47 ($+$0.15)
& 50.09 ($-$0.23)
& 50.01 ($-$0.31)
& 50.02 ($-$0.30)
& 49.95 ($-$0.37) \\
& Comp
& 0.76 & 0.78 & 0.81 & 0.83 & 0.85 & 0.87 \\
\midrule

\multirow{4}{*}{$\beta=0.5$}
& FA
& 0.28 & 0.28 & 0.28 & 0.72 & 0.44 & 0.28 \\
& RA
& 79.26 & 82.76 & 84.43 & 85.97 & 86.40 & 86.37 \\
& MIA AUC
& 49.99 ($-$0.33)
& 50.13 ($-$0.19)
& 50.04 ($-$0.28)
& 50.00 ($-$0.32)
& 50.13 ($-$0.19)
& 50.05 ($-$0.27) \\
& Comp
& 0.79 & 0.82 & 0.85 & 0.88 & 0.89 & 0.91 \\
\midrule

\multirow{4}{*}{$\beta=0.9$}
& FA
& 0.72 & 1.27 & 0.99 & 0.44 & 0.44 & 0.28 \\
& RA
& 84.24 & 85.49 & 85.99 & 86.89 & 86.99 & 86.76 \\
& MIA AUC
& 50.26 ($-$0.06)
& 50.22 ($-$0.10)
& 50.16 ($-$0.15)
& 49.91 ($-$0.41)
& 50.07 ($-$0.25)
& 50.05 ($-$0.27) \\
& Comp
& 0.81 & 0.85 & 0.87 & 0.89 & 0.91 & 0.92 \\
\midrule

\multicolumn{2}{c|}{Comm}
& 4.30 & 4.46 & 4.63 & 4.79 & 4.96 & 5.13 \\

\bottomrule
\end{tabular}}
\end{table*}

\begin{wraptable}{r}{0.5\textwidth}
\vspace{-1.5\intextsep}
\centering
\caption{Ablation of {\proposed} against its selection variants, with $\beta=0.5$, $\lambda=0.7$, and $k=25$ clients selected for unlearning. $\Delta$MIA is the signed deviation of MIA AUC from the retrained reference.}
\label{tab:ablation_proposed}
\resizebox{0.5\textwidth}{!}{%
\begin{tabular}{l|ccc|ccc}
\toprule
& \multicolumn{3}{c|}{$\alpha=0.1$} & \multicolumn{3}{c}{$\alpha=1.0$} \\
\cmidrule(lr){2-4} \cmidrule(lr){5-7}
Method & FA & RA & $\Delta$MIA & FA & RA & $\Delta$MIA \\
\midrule
\multicolumn{7}{c}{\textit{FMNIST-CNN} \; (Retrain MIA AUC: 50.32 / 50.23)} \\
\midrule
\proposed & 0.72 & \textbf{85.97} & $-$0.32 & \textbf{0.28} & \textbf{89.35} & \textbf{$+$0.02} \\
\nonclp   & \textbf{0.28} & 85.35 & $-$0.16 & 0.43 & 89.26 & \textbf{$+$0.02} \\
\reverse  & 1.36 & 82.28 & $+$0.04 & 1.33 & 89.14 & $-$0.06 \\
\cucoh    & 1.24 & 82.15 & $-$0.34 & 1.23 & 89.04 & $+$0.03 \\
\midrule
\multicolumn{7}{c}{\textit{CIFAR-10-ResNet} \; (Retrain MIA AUC: 55.64 / 52.18)} \\
\midrule
\proposed & \textbf{0.00} & \textbf{54.48} & $-$0.88 & \textbf{1.09} & \textbf{82.84} & $-$0.40 \\
\nonclp   & 0.19 & 49.60 & $-$1.10 & 1.51 & 81.51 & $+$0.12 \\
\reverse  & 4.44 & 41.09 & $-$0.77 & 1.30 & 81.91 & $+$0.15 \\
\cucoh    & 7.65 & 37.65 & $-$1.28 & 1.20 & 81.74 & $+$0.18 \\
\bottomrule
\end{tabular}}

\end{wraptable}
\textbf{Variants of {\proposed}.} Table~\ref{tab:ablation_proposed} evaluates the contribution of the proposed client-selection rule while keeping the unlearning procedure and number of selected clients fixed. Under highly heterogeneous data ($\alpha=0.1$), {\proposed} consistently achieves higher RA than {\reverse}, {\cucoh}, and {\nonclp}. Compared with {\reverse}, which also use $\pdi_i(\Itar)$ for client selection but in a reverse order, it improves RA by $3.69$ percentage points on FMNIST and $13.39$ points on CIFAR-10, while reducing FA by $0.64$ and $4.44$ points, respectively. Selecting highly target-coherent clients ({\cucoh}) further degrades both FA and RA, particularly on CIFAR-10, where FA increases from $0\%$ to $7.65\%$ and RA decreases from $54.48\%$ to $37.65\%$. Removing {\clp} weighting ({\nonclp}) also reduces RA, by $0.62$ points on FMNIST and $4.88$ points on CIFAR-10, showing that emphasizing critical rounds improves client selection. Under $\alpha=1.0$, the differences are smaller, although {\proposed} retains the highest RA on both datasets. Across all variants, $|\Delta\mathrm{MIA}|$ remains at most $1.28$ percentage points from Retrain. \textbf{These results show that the proposed ranking direction and {\clp}-aware history contribute to selecting retained clients that preserve utility while maintaining effective forgetting, supporting our central hypothesis.}

\section{Conclusion} We revisit client-level FU from a path-dependent perspective, motivated by the insight that retained clients are not equally valuable for unlearning: their usefulness depends on when and how their influence was accumulated during training and how strongly it is coupled with the target clients. Based on this insight, {\proposed} combines \clp-weighted historical global-model alignment and client-target coherence to rank retained clients, and performs retraining-based unlearning using only a selected top-$k$ subset. Experiment results demonstrate that {\proposed} achieves effective forgetting and strong retained utility while substantially reducing client participation, communication, and computation, and can be adapted to different application requirements by adjusting the forgetting–utility trade-off.

\begin{ack}
This publication has emanated from research conducted with the financial support of Research Ireland under Grant number 18/CRT/6222; School of Computer Science and Statistics, Trinity College Dublin. The work of Quoc-Viet Pham is supported in part by Research Ireland under the European Innovation Council CHIST-ERA SHIELD project (Project No. 216449, Award No. 19226). For the purpose of Open Access, the author has applied a CC BY public copyright licence to any Author Accepted Manuscript version arising from this submission.
\end{ack}


\bibliographystyle{plainnat}
\bibliography{ref}

@inproceedings{zhong2025unlearning,
  title={Unlearning through knowledge overwriting: Reversible federated unlearning via selective sparse adapter},
  author={Zhong, Zhengyi and Bao, Weidong and Wang, Ji and Zhang, Shuai and Zhou, Jingxuan and Lyu, Lingjuan and Lim, Wei Yang Bryan},
  booktitle={Proceedings of the Computer Vision and Pattern Recognition Conference},
  pages={30661--30670},
  year={2025}
}

@ARTICLE{11359714,
author={Nguyen, Thanh Linh and Oliveira, Marcela Tuler De and Braeken, An and Ding, Aaron Yi and Pham, Quoc-Viet},
journal={ IEEE Internet Computing },
title={{Toward Verifiable Federated Unlearning: Framework, Challenges, and the Road Ahead}},
year={2026},
volume={30},
number={01},
pages={59-70},
}

@article{gu2024ferrari,
  title={Ferrari: Federated feature unlearning via optimizing feature sensitivity},
  author={Gu, Hanlin and Ong, Win K and Chan, Chee S and Fan, Lixin},
  journal={Advances in Neural Information Processing Systems},
  volume={37},
  pages={24150--24180},
  year={2024}
}

@inproceedings{shokri2017membership,
  title={Membership inference attacks against machine learning models},
  author={Shokri, Reza and Stronati, Marco and Song, Congzheng and Shmatikov, Vitaly},
  booktitle={IEEE Symposium on Security and Privacy},
  pages={3--18},
  year={2017},
}

@inproceedings{yan2023criticalfl,
  title={Criticalfl: A critical learning periods augmented client selection framework for efficient federated learning},
  author={Yan, Gang and Wang, Hao and Yuan, Xu and Li, Jian},
  booktitle={Proceedings of the 29th ACM SIGKDD Conference on Knowledge Discovery and Data Mining},
  pages={2898--2907},
  year={2023}
}

@inproceedings{huang2025pa3fed,
  title={Pa3fed: Period-aware adaptive aggregation for improved federated learning},
  author={Huang, Chengxiang and Liu, Bingyan},
  booktitle={Proceedings of the AAAI Conference on Artificial Intelligence},
  volume={39},
  pages={17395--17403},
  year={2025}
}

@article{chundawat2024conda,
  title={Conda: Fast federated unlearning with contribution dampening},
  author={Chundawat, Vikram S and Niroula, Pushkar and Dhungana, Prasanna and Schoepf, Stefan and Mandal, Murari and Brintrup, Alexandra},
  journal={arXiv preprint arXiv:2410.04144},
  year={2024}
}

@inproceedings{foster2024fast,
  title={Fast machine unlearning without retraining through selective synaptic dampening},
  author={Foster, Jack and Schoepf, Stefan and Brintrup, Alexandra},
  booktitle={Proceedings of the AAAI conference on artificial intelligence},
  volume={38},
  number={11},
  pages={12043--12051},
  year={2024}
}

@inproceedings{wang2025fedcm,
  title={{FedCM}: Client clustering and migration in federated learning via gradient path similarity and update direction deviation},
  author={Wang, Peng and Lu, Shoupeng and Yin, Hao and Yang, Banglie and Zhu, Tianli and Dai, Cheng},
  booktitle={Proceedings of the Thirty-Fourth International Joint Conference on Artificial Intelligence},
  pages={6343--6351},
  year={2025}
}

@inproceedings{
oh2022fedbabu,
title={Fed{BABU}: Toward Enhanced Representation for Federated Image Classification},
author={Jaehoon Oh and SangMook Kim and Se-Young Yun},
booktitle={International Conference on Learning Representations},
year={2022},
}

@inproceedings{yi2024fedssa,
  title={{FedSSA}: semantic similarity-based aggregation for efficient model-heterogeneous personalized federated learning},
  author={Yi, Liping and Yu, Han and Shi, Zhuan and Wang, Gang and Liu, Xiaoguang and Cui, Lizhen and Li, Xiaoxiao},
  booktitle={Proceedings of the Thirty-Third International Joint Conference on Artificial Intelligence},
  pages={5371--5379},
  year={2024}
}

@inproceedings{li2023no,
  title={No Fear of Classifier Biases: Neural collapse inspired federated learning with synthetic and fixed classifier},
  author={Li, Zexi and Shang, Xinyi and He, Rui and Lin, Tao and Wu, Chao},
  booktitle={Proceedings of the IEEE/CVF International Conference on Computer Vision},
  pages={5319--5329},
  year={2023}
}

@article{luo2021no,
  title={{No Fear of Heterogeneity}: Classifier calibration for federated learning with Non-IID data},
  author={Luo, Mi and Chen, Fei and Hu, Dapeng and Zhang, Yifan and Liang, Jian and Feng, Jiashi},
  journal={Advances in Neural Information Processing Systems},
  volume={34},
  pages={5972--5984},
  year={2021}
}

@inproceedings{liu2021federaser,
  title={Federaser: Enabling efficient client-level data removal from federated learning models},
  author={Liu, Gaoyang and Ma, Xiaoqiang and Yang, Yang and Wang, Chen and Liu, Jiangchuan},
  booktitle={2021 IEEE/ACM 29th international symposium on quality of service (IWQOS)},
  pages={1--10},
  year={2021}
}

@inproceedings{xiong2023exact,
  title={{Exact-Fun}: An exact and efficient federated unlearning approach},
  author={Xiong, Zuobin and Li, Wei and Li, Yingshu and Cai, Zhipeng},
  booktitle={2023 IEEE International Conference on Data Mining (ICDM)},
  pages={1439--1444},
  year={2023}
}

@inproceedings{cao2015towards,
  title={Towards making systems forget with machine unlearning},
  author={Cao, Yinzhi and Yang, Junfeng},
  booktitle={2015 IEEE Symposium on Security and Privacy},
  pages={463--480},
  year={2015}
}

@article{jia2023model,
  title={Model sparsity can simplify machine unlearning},
  author={Jia, Jinghan and Liu, Jiancheng and Ram, Parikshit and Yao, Yuguang and Liu, Gaowen and Liu, Yang and Sharma, Pranay and Liu, Sijia},
  journal={Advances in Neural Information Processing Systems},
  volume={36},
  pages={51584--51605},
  year={2023}
}

@inproceedings{golatkar2020eternal,
  title={Eternal sunshine of the spotless net: Selective forgetting in deep networks},
  author={Golatkar, Aditya and Achille, Alessandro and Soatto, Stefano},
  booktitle={Proceedings of the IEEE/CVF Conference on Computer Vision and Pattern Recognition},
  pages={9304--9312},
  year={2020}
}

@inproceedings{brophy2021machine,
  title={Machine unlearning for random forests},
  author={Brophy, Jonathan and Lowd, Daniel},
  booktitle={International Conference on Machine Learning},
  pages={1092--1104},
  year={2021}
}

@article{romandini2024federated,
  title={Federated unlearning: A survey on methods, design guidelines, and evaluation metrics},
  author={Romandini, Nicol{\`o} and Mora, Alessio and Mazzocca, Carlo and Montanari, Rebecca and Bellavista, Paolo},
  journal={IEEE Transactions on Neural Networks and Learning Systems},
  volume={36},
  number={7},
  pages={11697--11717},
  year={2024}
}

@inproceedings{mcmahan2017communication,
  title={Communication-efficient learning of deep networks from decentralized data},
  author={McMahan, Brendan and Moore, Eider and Ramage, Daniel and Hampson, Seth and y Arcas, Blaise Aguera},
  booktitle={Artificial Intelligence and Statistics},
  pages={1273--1282},
  year={2017},
}

@ARTICLE{11415662,
  author={Jeong, Hyejun and Ma, Shiqing and Houmansadr, Amir},
  journal={IEEE Transactions on Big Data}, 
  title={Survey on Federated Unlearning: Challenges and Opportunities}, 
  year={2026},
  volume={12},
  number={3},
  pages={702-720},}

@article{zhang2023fedrecovery,
  title={FedRecovery: Differentially private machine unlearning for federated learning frameworks},
  author={Zhang, Lefeng and Zhu, Tianqing and Zhang, Haibin and Xiong, Ping and Zhou, Wanlei},
  journal={IEEE Transactions on Information Forensics and Security},
  volume={18},
  pages={4732--4746},
  year={2023}
}

@inproceedings{khalil2025not,
  title={Not: Federated unlearning via weight negation},
  author={Khalil, Yasser H and Brunswic, Leo and Lamghari, Soufiane and Li, Xu and Beitollahi, Mahdi and Chen, Xi},
  booktitle={Proceedings of the Computer Vision and Pattern Recognition Conference},
  pages={25759--25769},
  year={2025}
}

@inproceedings{gong2026fedortho,
  title={{FedOrtho}: Efficient Federated Unlearning Via Orthogonal Convolution and Adaptive Soft Pruning},
  author={Gong, Qinghui and Yang, Xue and Chen, Xunlei and Lai, Jinshan and Meng, Hua and Tang, Xiaohu},
  booktitle={Proceedings of the IEEE/CVF Conference on Computer Vision and Pattern Recognition},
  pages={8009--8018},
  year={2026}
}

@ARTICLE{10810730,
  author={Wang, Weiqi and Zhang, Chenhan and Tian, Zhiyi and Yu, Shui},
  journal={IEEE Transactions on Dependable and Secure Computing}, 
  title={{FedU}: Federated Unlearning via User-Side Influence Approximation Forgetting}, 
  year={2025},
  volume={22},
  number={3},
  pages={2550-2562}}

@inproceedings{pan2025federated,
  title={Federated unlearning with gradient descent and conflict mitigation},
  author={Pan, Zibin and Wang, Zhichao and Li, Chi and Zheng, Kaiyan and Wang, Boqi and Tang, Xiaoying and Zhao, Junhua},
  booktitle={Proceedings of the AAAI Conference on Artificial Intelligence},
  volume={39},
  number={19},
  pages={19804--19812},
  year={2025}
}

@inproceedings{nguyen2026computation,
  title={Computation and Communication Efficient Federated Unlearning via On-server Gradient Conflict Mitigation and Expression},
  author={Nguyen, Minh-Duong and Wanasekara, Senura and Nguyen, Le-Tuan and Pham, Quoc-Viet and Yong, Ken-Tye and Tran, Nguyen H and Le, Dung D},
  booktitle={Proceedings of the IEEE/CVF Conference on Computer Vision and Pattern Recognition},
  pages={3347--3357},
  year={2026}
}

@inproceedings{achille2018critical,
  title={Critical learning periods in deep networks},
  author={Achille, Alessandro and Rovere, Matteo and Soatto, Stefano},
  booktitle={International Conference on Learning Representations},
  year={2018}
}

@inproceedings{kleinman2024critical,
      title={{Critical learning periods emerge even in deep linear networks}}, 
      author={Michael Kleinman and Alessandro Achille and Stefano Soatto},
      booktitle={International Conference on Learning Representations}, 
      year={2024},
}

@inproceedings{yan2022seizing,
  title={Seizing critical learning periods in federated learning},
  author={Yan, Gang and Wang, Hao and Li, Jian},
  booktitle={Proceedings of the AAAI Conference on Artificial Intelligence},
  volume={36},
  issue={8}, 
  pages={8788--8796},
  year={2022}
}

@inproceedings{wang2026gr,
  title={{GR-Gauge}: Cost-efficient Training Configuration By Gauging the Gradient Redundancy},
  author={Wang, Guanjie and Chen, Chen},
  booktitle={Proceedings of the IEEE/CVF Conference on Computer Vision and Pattern Recognition},
  pages={12934--12943},
  year={2026}
}

@misc{gdpr,
  author = {European Parliament and Council of the European Union},
  title = {{General Data Protection Regulation}},
  howpublished = {[Online]. Available: "\url{https://gdpr-info.eu/}"},
  year = {2016}, 
  note = "(Accessed Jun. 07, 2026)"
}

@article{antonelli1997economics,
  title={The economics of path-dependence in industrial organization},
  author={Antonelli, Cristiano},
  journal={International Journal of Industrial Organization},
  volume={15},
  number={6},
  pages={643--675},
  year={1997},
  publisher={Elsevier}
}

@article{xiao2017fashion,
  title={{Fashion-MNIST}: A novel image dataset for benchmarking machine learning algorithms},
  author={Xiao, Han and Rasul, Kashif and Vollgraf, Roland},
  journal={arXiv:1708.07747},
  year={2017}
}

@inproceedings{weng2026gdfa,
  title={{GDFA}: Geometry-Driven Federated Unlearning with Directional Task Vector Alignment},
  author={Weng, Xiuting and Pu, Ruizhi and Yao, Yuanhang and Yue, Kun and Tang, Zhiwen and Yu, Lixing},
  booktitle={Proceedings of the IEEE/CVF Conference on Computer Vision and Pattern Recognition},
  pages={10346--10356},
  year={2026}
}

@article{hanser2025data,
  title={Data-driven federated learning in drug discovery with knowledge distillation},
  author={Hanser, Thierry and Ahlberg, Ernst and Amberg, Alexander and Anger, Lennart T and Barber, Chris and Brennan, Richard J and Brigo, Alessandro and Delaunois, Annie and Glowienke, Susanne and Greene, Nigel and others},
  journal={Nature Machine Intelligence},
  volume={7},
  number={3},
  pages={423--436},
  year={2025},
  publisher={Nature Publishing Group UK London}
}

@article{theophanous2026international,
  title={An international multi-centre study to develop and validate federated learning-based prognostic models for anal cancer},
  author={Theophanous, Stelios and L{\o}nne, Per-Ivar and Choudhury, Ananya and Berbee, Maaike and Deijen, Charlotte and Dekker, Andre and Field, Matthew and Gambacorta, Maria Antonietta and Gilbert, Alexandra and Gr{\o}nlie Guren, Marianne and others},
  journal={Nature Communications},
  volume={17},
  number={1},
  pages={3956},
  year={2026},
  publisher={Nature Publishing Group UK London}
}

@misc{cifar,
  title        = "The CIFAR-10 and CIFAR-100 datasets",
  author       = "{Alex Krizhevsky, Vinod Nair, and Geoffrey Hinton}",
  howpublished = "\url{https://cave.cs.toronto.edu/kriz/cifar.html}",
  year         = 2009,
  note         = "Accessed: 2026-06-18"
}

@inproceedings{he2016deep,
  title={Deep residual learning for image recognition},
  author={He, Kaiming and Zhang, Xiangyu and Ren, Shaoqing and Sun, Jian},
  booktitle={Proceedings of the IEEE conference on computer vision and pattern recognition},
  pages={770--778},
  year={2016}
}


\appendix
\section{Details of Experimental Evaluation}

\subsection{CNN Architecture for FMNIST}
\label{subsec:cnn}
The CNN architecture used for FMNIST consists of two modules: a feature extractor and a classifier. The feature extractor comprises three convolutional blocks. The first block applies a \texttt{Conv2d} layer with 1 input channel and 32 output channels (kernel size 3), followed by Batch Normalization, ReLU activation, and a \texttt{Dropout2d} layer (rate 0.3). The second block applies a \texttt{Conv2d} layer with 32 input channels and 64 output channels (kernel size 3, stride 1), followed by Batch Normalization, ReLU activation, and a $2\times2$ max-pooling layer. The third block applies a \texttt{Conv2d} layer with 64 input and output channels (kernel size 3), followed by Batch Normalization, ReLU activation, a $4\times4$ max-pooling layer, and a \texttt{Dropout2d} layer (rate 0.2). The resulting feature map of size $64\times3\times3$ is flattened into a 576-dimensional vector. The classifier consists of a fully connected layer ($576\rightarrow256$) with ReLU activation and Dropout (rate 0.1), followed by a final linear layer ($256\rightarrow10$). All convolutional and linear layers are initialized with Kaiming uniform initialization with zero biases.

\subsection{Federated Unlearning Baselines}
\label{subsec:baseline}
Below are details of state-of-the-art FU approaches which are used to evaluate the efficacy and effectiveness of {\proposed}.
\begin{itemize}
    \item \textbf{Retrain}. As the gold-standard unlearning baseline, Retrain removes the target clients entirely and trains from scratch on the remaining clients. While this guarantees exact unlearning, it is prohibitively expensive in practice due to substantial computational overhead and reliance on the participation availability of all remaining clients.
    \item \textbf{FedEraser}~\cite{liu2021federaser}. FedEraser approximates the retrained model without incurring its full cost by replaying periodic historical parameter updates cached from remaining clients through a calibrated aggregation procedure, achieving significant reductions in both time and storage overhead relative to full retraining.
    \item \textbf{Exact-Fun} (E-F)~\cite{xiong2023exact}. Exact-Fun (E-F) leverages quantized FL to stabilize the global model, enabling exact unlearning with high efficiency. Upon a deletion request, only the affected client retrains locally. Retraining of the global model is triggered solely when the quantized model state changes, allowing many requests to be resolved with minimal overhead.
    \item \textbf{CONDA}~\cite{chundawat2024conda, foster2024fast}. CONDA is a server-side unlearning method that eliminates a target client's influence without client participation. It tracks per-client parameter contributions throughout training, estimates the forgotten client's relative influence across model parameters, and applies a targeted dampening factor to parameters most strongly associated with the forgotten client.

    \item \textbf{FedRecovery}~\cite{zhang2023fedrecovery}. FedRecovery is a server-side unlearning method that does not require retraining-based calibration and fine-tuning. It removes the target client’s accumulated gradient residuals from the trained global model cached from the learning phase. It also injects calibrated Gaussian noise to mask the remaining distance between the recovered model and the ideal retrained model, thereby making the two models statistically indistinguishable.
    
    \item \textbf{FUSED}~\cite{zhong2025unlearning}. FUSED confines unlearning to the most task-critical layers identified during training, attaching lightweight sparse adapters exclusively to these layers and fine-tuning them on remaining clients to overwrite the forgotten knowledge. This design substantially reduces unlearning cost while preserving reversibility.

    \item \textbf{FOUL}~\cite{nguyen2026computation}. FOUL is an on-server unlearning method that first disentangles the model into causal and non-causal components during the learning-to-unlearn stage, and then performs unlearning on the domain-specific non-causal component. In the unlearning stage, FOUL constructs an aggregated unlearning gradient by optimizing learnable coefficients over client gradients, encouraging alignment with retain-client gradients while reducing alignment with target-client gradients. This design preserves retain-client knowledge while removing forget-client-specific influence without requiring the server to access client data.

    \item \textbf{GDFA}~\cite{weng2026gdfa}. GDFA is a model-manipulation-based federated unlearning method that exploits loss-landscape geometry and task-vector alignment to improve unlearning under data heterogeneity. Upon an unlearning request, GDFA constructs multiple perturbed models around the trained global model and fine-tunes them on the forgetting data to derive task vectors representing the target knowledge; directionally consistent components are then merged through sign consensus and negated from the global model. By operating around flatter loss regions, GDFA aims to improve robustness to parameter modification and preserve retained knowledge, but requires request-time access to the forgetting data and incurs additional computation from generating and fine-tuning multiple perturbed models.

\end{itemize}



\end{document}